\documentclass[runningheads]{llncs}
\usepackage[T1]{fontenc}
\usepackage{amsmath}
\usepackage{amsfonts}
\usepackage{algorithmicx}
\usepackage{textcomp}
\usepackage{xcolor}
\usepackage{todonotes}
\setuptodonotes{inline}
\usepackage{graphicx}
\graphicspath{{./images/}}
\usepackage[english]{babel}
\usepackage{csquotes}
\usepackage[backend=biber, natbib=true, style=ieee]{biblatex} 
\usepackage[hidelinks]{hyperref}
\hypersetup{colorlinks=false, pdftitle={Towards Resolving Word Ambiguity with Word Embeddings}}
\usepackage{makecell}
\usepackage{tikz}
\usetikzlibrary{mindmap}

\usepackage{booktabs, multirow, tabularx}
\newcolumntype{C}{>{\centering\arraybackslash}X}
\NewExpandableDocumentCommand\mcc{O{1}m}{\multicolumn{#1}{c}{#2}}
\usepackage{setspace}
\usepackage[skip=1ex, font={footnotesize, sf, stretch=0.88}, labelfont=bf]{caption}

\begin{document}

    \title{Towards Resolving Word Ambiguity with \\ Word Embeddings\thanks{This research is funded by the State of Bavaria.}}

    \author{
        Matthias Thurnbauer\inst{1} \and
        Johannes Reisinger\inst{2}\orcidID{0009-0009-5844-795X} \and
        Christoph Goller\inst{1} \and
        Andreas Fischer\inst{2}\orcidID{0000-0002-0074-5411}
    }

    \authorrunning{Thurnbauer et al.}

    \institute{
        IntraFind Software AG, Landsberger Str. 368, 80687 Munich, Germany
        \url{https://intrafind.de} \\
        \email{thurnbauermatthias@gmail.com \\ christoph.goller@intrafind.de} \and
        Deggendorf Institute of Technology, Dieter-Görlitz-Platz 1, 94469 Deggendorf, Germany
        \url{https://www.th-deg.de/} \\
        \email{\{johannes.reisinger, andreas.fischer\}@th-deg.de}
    }

    \maketitle

    \begin{abstract}
    Ambiguity is ubiquitous in natural language.
    Resolving ambiguous meanings is especially important in information retrieval tasks.
    While word embeddings carry semantic information, they fail to handle ambiguity well.
    Transformer models have been shown to handle word ambiguity for complex queries, but they cannot be used to
    identify ambiguous words, e.g. for a 1-word query.
    Furthermore, training these models is costly in terms of time, hardware resources, and training data,
    prohibiting their use in specialized environments with sensitive data.
    Word embeddings can be trained using moderate hardware resources.
    This paper shows that applying DBSCAN clustering to the latent space can identify ambiguous words and
    evaluate their level of ambiguity.
    An automatic DBSCAN parameter selection leads to high-quality clusters, which are semantically coherent and
    correspond well to the perceived meanings of a given word.
\end{abstract}


    \section{Introduction}\label{sec:introduction}
Many words are ambiguous.
The English word \textit{bank} may denote a financial institution or a riverside.
Besides such well-known ambiguities, it turns out that most words are ambiguous, at least to a certain extent.
For example the word \textit{Coca-Cola} may refer to a specific soft-drink, the respective company, as well as the
shares of that company.

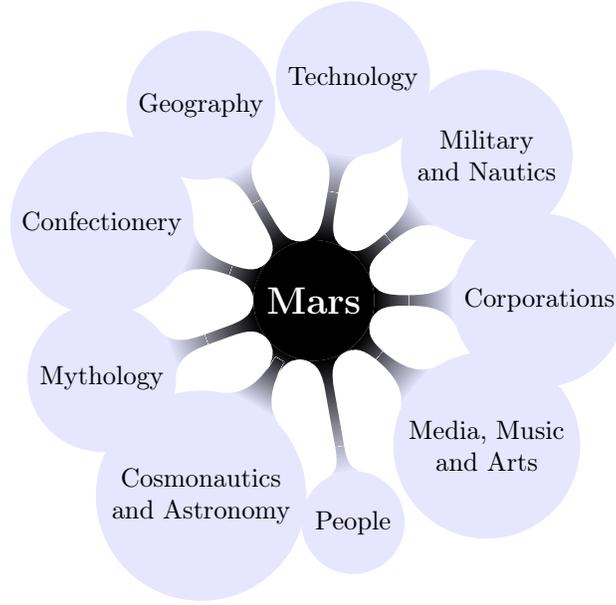
\begin{figure}[htbp]
    \centering
    \begin{tikzpicture}[
        grow cyclic,
        align=flush center,
        every node/.style=concept,
        concept color=blue!10,
        level 1/.style={level distance=3cm,sibling angle=40}]
        \node[concept color=black,text=white] {\Large \textbf{Mars}}
        child { node {Mythology}} 
        child { node {Cosmonautics \\ and Astronomy}} 
        child { node {People}} 
        child { node {Media, Music \\ and Arts}} 
        child { node {Corporations}} 
        child { node {Military \\ and Nautics}} 
        child { node {Technology}} 
        child { node {Geography}} 
        child { node {Confectionery}} 
        ;
    \end{tikzpicture}
    \caption{The 9 main domains for the word ``Mars'' according to the wikipedia ambiguity page,
        containing a total of 86 wikipedia pages of homonymous words~\cite{Mars2023}.}
    \label{fig:mars_meanings}
\end{figure}

Ambiguities are often related to topics or domains.
Within the domain \textit{finance and economy}, the word \textit{bank} almost always denotes a financial institution
rather than a riverside.
However, within this domain there may be further ambiguities of the word \textit{bank}.
Depending on the document context, \textit{bank} may refer to multiple individual financial institutions.
Figure~\ref{fig:mars_meanings} illustrates various domains for the word \textit{Mars},
each bearing multiple meanings.

Shifting the context of an ambiguous word also changes its meaning, leading to misunderstanding, and
creating the necessity to reaffirm the context or ask for further information.
These communication phenomena occur in both human-to-human and human-to-machine interaction.
A search engine or a chatbot might, for example, have to ask for further information before answering an ambiguous query
or question.
While this implies the necessity to recognize ambiguities in queries,
we are less interested in identifying and resolving general ambiguities.
Instead, we focus on ambiguities w.r.t.\ a specific data set (document corpus) and knowledge available to the system,
since these are the ambiguities most relevant for query/question answering.

In his \textit{Philosophical Investigations}~\cite{wittgenstein1953philosophical} Wittgenstein claimed
\textit{`The meaning of a word is its use in the language'} (§43).
We translate \textit{`use in language'} to context.
The context may contain information about the interacting participants, location, past interactions, as well as non-verbal
communication.
Within this paper we focus on textual context of a target word, more precisely,
the five tokens preceding and following a target word.
This quote of Wittgenstein can be seen as the philosophical foundation of current approaches to
Natural Language Processing.

Transformers~\cite{vaswani2017attention} produce embeddings for sentences or paragraphs.
The representations they produce for individual words are highly context dependent.
However, transformers do not help us identify or resolve ambiguity.
A generative language model such as ChatGPT may produce a reasonable answer for the question
\textit{`What are the different meanings of the word bank'}, but it cannot be easily used to identify ambiguities of
a word w.r.t.\ a given document corpus~\cite{ChatGPT}.
We decided to look into much simpler word embedding models since they require much less resources
to train and operate and results are easier to understand and interpret.
Word embedding models produce one embedding per word based on all contexts of the training corpus (document corpus),
which makes them more compatible with the bag-of-words model, the foundation of classical search engine technology.
The focus of this paper is to identify and resolve word ambiguity in unsupervised manner, using the CBOW
(continuous bag of words) architecture from word2vec~\cite{mikolov2013efficient} conjoined with the
DBSCAN~\cite{DensitybasedAlgorithmDiscovering1996} clustering algorithm.

We first show similarities and differences to related work in section~\ref{sec:related-work}.
Section~\ref{sec:data-and-preprocessing} contains data selection, preprocessing and tokenization,
as well as the training of word embeddings and the generation of context vectors.
We then describe the clustering process and the proposed parameter score for automatic cluster parameter
selection in section~\ref{sec:clustering-and-automatic-parameter-selection}.
Finally, we qualitatively analyse the clustering results and discuss next steps.
The code used in this paper is freely accessible at\newline
\url{https://github.com/thurnbauermatthi/WordDisambiguation}.

    \section{Related Work}\label{sec:related-work}

\subsection{Word2Vec and Dual Embedding Space}\label{subsec:word2vec}
\citeauthor{mitraDualEmbeddingSpace2016}~\cite{mitraDualEmbeddingSpace2016} focus on the role of the two different
embedding spaces in Word2Vec~\cite{mikolov2013efficient} models.
Word2Vec calculates two different weight matrices with dimension $V \times D$, where $V$ is the vocabulary size and $D$
is the feature (embedding) size.
For the CBOW model, the embeddings in the output space (OUT) are optimized towards a large inner-product with the
mean of their context word vectors in the input space (IN).
After all the goal is to predict the center word based on its context words.
The authors show that the embeddings from these two spaces are dissimilar.
For the IN-to-OUT case, the most similar words to a given target word are words typically occurring in its context,
resulting in thematic similarity.
For example, the resulting top 2 words for the target word ``eminem'' are ``rap'' and ``featuring''.
But for the IN-to-IN and OUT-to-OUT case, the most similar words to a given target word are similar in type and function.
Here, the resulting top 2 words for the target word ``eminem'' are ``rihanna'' and ``ludacris'',
and ``rihanna'' and ``dre'', respectively.
These words are similar, since they occur in similar contexts.
These insights on the dual embedding space help us to reflect our final results.

\subsection{Word Sense Disambiguation}\label{subsec:wsd}
Most work on word sense disambiguation (WSD) is focused on supervised methods.
There is either a training corpus for different meanings of ambiguous words, or an external ressource such as
WordNet~\cite{WordNetElectronicLexical1998} to provide examples of different meanings of ambiguous words.
For the latter, actual contexts of ambiguous words are compared to the contexts of these examples to disambiguate
the word~\cite{orkpholWordSenseDisambiguation2019, nurifanDevelopingCorporaUsing2018}.
\citeauthor{kutuzovLemmatizeNotLemmatize2019}~\cite{kutuzovLemmatizeNotLemmatize2019} work with
\textit{ELMo}~\cite{petersDeepContextualizedWord2018} networks.
As with other embedding architectures, lemmatisation (token normalization) is unnecessary and usually not applied.
However, their experiments show that for morphology-rich languages like Russian, lemmatized training data yields
improvements for WSD.
\citeauthor{alkhatlanWordSenseDisambiguation2018}~\cite{alkhatlanWordSenseDisambiguation2018} utilise
Word2Vec and GloVe to disambiguate polysemous arabic words by using an Arabic version of WordNet,
as well as stemming to normalize the tokens.
Since we work with a German corpus and since German is also a morphology-rich language, we lemmatize the words
before training.

In contrast to most work on WSD our method is unsupervised and does not require an external database or labelled data.
We cluster the context vectors of potentially ambiguous words similar to~\cite{huangImprovingWordRepresentations2012}
and~\cite{schutze-1998-automatic}.
\citeauthor{huangImprovingWordRepresentations2012}~\cite{huangImprovingWordRepresentations2012} use a special network
architecture with both global and local context to produce word embeddings.
\citeauthor{schutze-1998-automatic}~\cite{schutze-1998-automatic} uses sparse vector representations for words.
Both authors make no attempt to automatically identify ambiguous words, but apply context-clustering with a fixed
number of clusters to all words or to known, ambiguous words, respectively.
However, we are investigating whether the results of clustering can be used to decide whether a word is ambiguous
and how ambiguous it is.

    \section{Data and Preprocessing}\label{sec:data-and-preprocessing}
This section contains data selection, preprocessing, word embedding training, and context vector generation.

\subsection{Data Selection}\label{subsec:data-selection}
We decided to use the wikimedia dump of all german wikipedia pages, containing
2\,724\,305 documents~\cite{WikimediaDownloads}.
Wikipedia summarizes their known ambiguities of words through a generic url,
allowing us to manually compare our results with this reference.
The utilized fields of the wikipedia dump are \textit{text} and \textit{auxiliary\_text},
disregarding the \textit{title} field, as it carries little information word embeddings training.
The \textit{text} field contains the body of the Wikipedia document, while the \textit{auxiliary\_text} field includes
thumbnail captions, tables, and other things which are not part of the \textit{text} field.
No disambiguation information is used for training as we focus on an unsupervised method.

\subsection{Preprocessing \& Tokenization}\label{subsec:tokenization}
We decided to use spaCy~\cite{spaCy} to tokenize and lemmatize our German text content since the SpaCy tokenization
pipeline achieved the most promising accuracy for tokenization and part-of-speech tagging,
compared to other state-of-the-art NLP libraries like Google's SyntaxNet, Stanford's CoreNLP,
and NLTK~\cite{alomranChoosingNLPLibrary2017}.
We also used the built-in named entity recognition to generate compound tokens, and the medium-sized spaCy core news
model, which was trained on german news~\cite{SpacyCoreNews}.

\subsection{Word Embedding Training}\label{subsec:word-embedding-training}
We used the Word2Vec~\cite{mikolov2013efficient} implementation of the Gensim~\cite{gensim} library to train word
embeddings with the CBOW model.
The word embeddings were created by utilizing the multiprocess functionality.
We used a \textit{window size} of 5, 12 \textit{worker threads}, a \textit{min count} of 10 and trained for
5 \textit{epochs}.
Training took approximately 40 minutes on a workstation with a 12 core Intel(R) Xeon(R) W-2235 CPU, 128 GB RAM,
with no GPUs used.

\subsection{Context Vector Generation}\label{subsec:context-vector-generation}
In a CBOW network a window of words/tokens (e.g 5 tokens to the left and 5 tokens to the right) defines
the context of a target word for each occurrence of this target word in the text corpus.
For each occurrence the embeddings of the context words (IN layer of the network) are summed up to get a context vector.
Table~\ref{tab:word-embeddings} shows the number of context vectors for some test words.
The context vectors for each word were generated in about 145 seconds to 155 seconds on a 12-core Intel i7-12700 CPU.
We manually counted the number of domains and Wikipedia Pages of Homonyms, to then manually evaluate the clustering
results.

\begin{table}[h]
    \centering
    \caption{Selected words with their number of manually counted domains, wikipedia pages of homonyms,
        and context vector information.}
    \label{tab:word-embeddings}
    \begin{tabular}{lrrrr}
        \toprule
        \thead{Word} & \thead{Domains} & \thead{Wikipedia Pages \\ of Homonyms} & \thead{Context Vectors} &
        \thead{Context Vector \\ File Size {[}kB{]}}  \\ \midrule
        Mars & 9 & 86 & 35899 & 14.359 \\
        \makecell[l]{Maus \\ (en. mouse)} & 14 & 51  & 28547  & 11.419    \\
        \makecell[l]{Kleeblatt \\ (en. cloverleaf)} & 6 & 14 & 4192 & 1.676 \\
        Pepsi & 2 & 4 & 2315 & 926 \\
        \makecell[l]{Gabelbein \\ (en. wishbone)} & 2 & 2 & 132 & 53 \\
        \makecell[l]{Datenstruktur \\ (en. data structure)} & 1 & 1 & 2592 & 1.036 \\
        \bottomrule
    \end{tabular}
\end{table}

Based on the assumption that the meaning of a word is defined by its context, we cluster the context
vectors of a target word to identify its meanings.
For each cluster, we can then compute a representative (centroid) embedding vector.
    \section{Clustering}\label{sec:clustering-and-automatic-parameter-selection}
In this section we describe our method to select clustering parameters which is derived with the goal
to create a similar number of clusters as there are meanings in Wikipedia.
We also describe our methods for deriving cluster labels.

\subsection{Clustering the Context Vectors}\label{subsec:clustering-the-context-vectors}
We chose DBSCAN~\cite{DensitybasedAlgorithmDiscovering1996}, a density-based clustering method, to retrieve both clearly
defined and easily interpretable clusters.
Each occurrence of a potentially ambiguous target word should be assigned to a single cluster based on its actual context.
The~\emph{noise cluster}, also generated by DBSCAN, contains those context vectors which are ambiguous in presence of
the target word and hence relate to multiple clusters.
These can not be assigned to a single cluster as their vectors lie in between multiple clusters.

DBSCAN uses two parameters to obtain its results.
Epsilon $\epsilon$ is the radius that defines the equidistant neighborhood of every data point,
while the minimum samples parameter $m$ defines the number of points required within the neighborhood of a given
data point, for it to be labeled \emph{core point}.
Data points with at least 1 neighbouring point, but fewer data points in their neighbourhood than $m$ are labeled
\emph{border point}.
Clusters are then generated from core points and border points.
Outlier points, which do not contain other points within their radius $\epsilon$, are assigned to the noise cluster through
the label ``-1''.

We clustered the context vectors for all combinations of the minimum sample sizes $m = \{ 5,10,20,30, \dots, 100 \}$
and $\epsilon = \{ 0.01, 0.02, \dots, 0.99 \}$.
The lowest minimum sample size $m$ is 5 to detect clusters that were generated from few-context words.
We do not set minPts to 2dim-1 as proposed in~\cite{schubertDBSCANRevisitedRevisited2017},
as we use cosine similarity rather than the Euclidean distance metric.
Following this heuristic would lead to clusterings where the numbers of clusters fails to reach the expected amount.

\begin{figure*}[h]
    \centering
    \includegraphics[width=\linewidth]{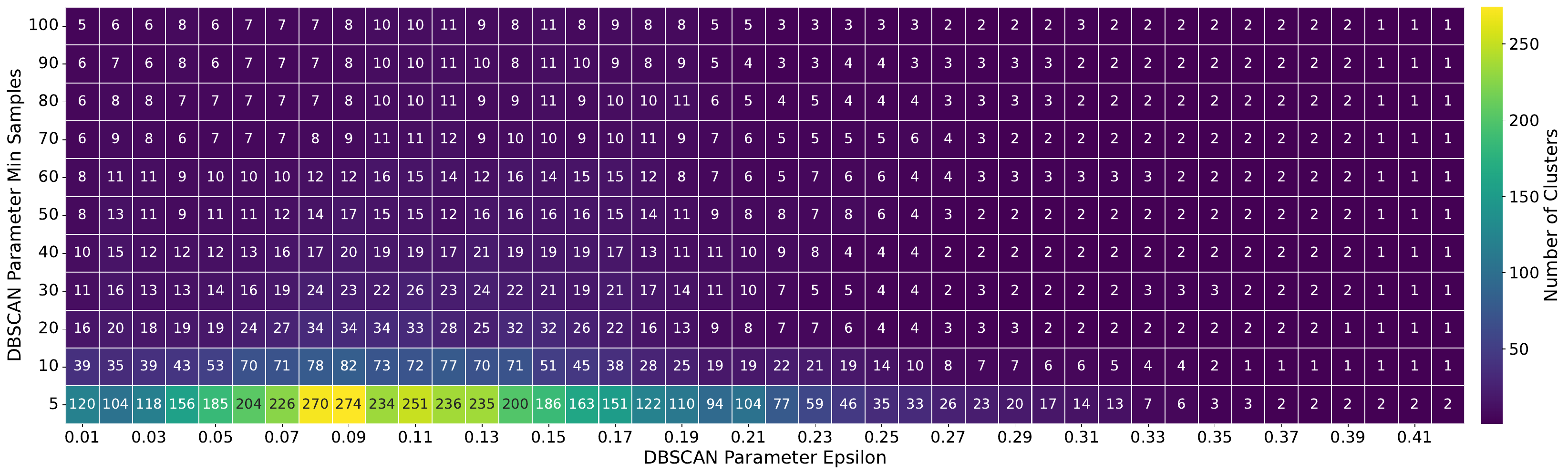}
    \caption{A heatmap of the number of clusters over a range of both the epsilon and minimum samples values
    for the word `mars'.
    The number of clusters for $0.43 \leq \epsilon \leq 1.00$ equal 1 and are not shown within this heatmap.}
    \label{fig:cluster_count}
\end{figure*}

Figure~\ref{fig:cluster_count} displays a section of the heatmap, visualizing the number of clusters for every
parameter combination for the word `mars'.
The heatmaps of all chosen words matched this heatmap pattern, where the highest number of clusters are in a range
of about $0.05 \leq \epsilon \leq 0.30$ and $5 \leq m \leq 20$, and the highest number of clusters occurring for $m = 5$.

\subsection{Silhouette Score Calculation}\label{subsec:silhouette-score-calculation}
To evaluate cluster quality, we calculated the silhouette score~\cite{rousseeuwSilhouettesGraphicalAid1987}
for every cluster, every word, and every parameter combination, using the respective sklearn
function~\cite{SklearnMetricsSilhouettescore}.
The silhouette score $s(c)$ of a cluster $c$ equals the difference between its mean intra-cluster
distance $a(c)$ and the mean inter-cluster distance $b(c)$ to its nearest cluster, w.r.t.\ the larger distance of
the two.
We use the cosine similarity metric instead of the Euclidean distance metric, analog to the clustering.

\begin{equation}
{s}(c)
    = \frac{b(c) - a(c)}{\max(a(c), b(c))}
    \label{eq:silhouette_score}
\end{equation}

Finally, the overall silhouette score $\overline{s}(C)$, of a clustering $C$ with $n(C)$ clusters,
is calculated as the mean of all silhouette scores ${s}(c)$, $c \in C$.

\subsection{Automatic DBSCAN Parameters Selection}\label{subsec:automatic-dbscan-parameters-selection}
\citeauthor{schubertDBSCANRevisitedRevisited2017} (\cite{schubertDBSCANRevisitedRevisited2017}) revisited common
heuristics, as well as `red flags for degenerate clustering results' used to determine DBSCAN parameters.
Following these heuristics and `red flags', we propose a formula to evaluate the results for a set of DBSCAN parameters.
The parameter score uses four clustering-specific variables: The number of clusters $n(c)$, the average
silhouette score $\overline{s}(c)$, the DBSCAN parameter $\epsilon$, and the noise ratio $noise(c)$.
Every parameter within formula~\ref{eq:dbscan-parameter-score} is normalized to ease interpretation of the
parameter score.

\begin{equation}
    p(c) = 2 \frac{\frac{n(c)}{max(n(c_i))} \cdot 0.5(\overline{s}(c) + 1)}
    {1 + (\frac{\epsilon(c)}{max(\epsilon(c_i))} \cdot \frac{noise(c)}{0.3})}
    \label{eq:dbscan-parameter-score}
\end{equation}

A high average silhouette score represents disjunctive clusters, and is simply shifted to the value range of $[0;1]$
for this equation.
Next, \citeauthor{schubertDBSCANRevisitedRevisited2017} advocate towards a lower epsilon value rather,
to preserve high cluster quality.
Dividing the number of data points in the noise cluster by the total number of data points equals the noise ratio.
As the desirable amount of noise is in the rage of 1\% to 30\%,
the parameter score is set to 0 when the noise score exceeds this range.
This cut is clearly visible in figure~\ref{fig:parameter-score}, where the noise ratio limits the parameter score
from the lower $\epsilon$ range, while the silhouette score limits the parameter score from upper $\epsilon$ range,
which equals 0 if only a single cluster except the noise cluster exists.
Both noise ratio and silhouette score are roughly proportional to the $\epsilon$ parameter,
but the number of clusters increases for lower to middle epsilon values and low minimum sample sizes.

\citeauthor{schubertDBSCANRevisitedRevisited2017} recommend to evaluate the relative size of clusters.
Yet, since we do not expect contexts to occur equally often in presence of the target word,
we do not apply this heuristic.

We chose to normalize all variables as well as the parameter score itself to a range of $[0;1]$ for eased interpretability.
This step is not mandatory when comparing parameters of the same set of vectors where only the clustering parameters
differ, but results in an interpretable score.
The number of clusters is divided by the highest cluster count, the epsilon value is divided by the highest epsilon
value that contains more than one cluster, except the noise cluster, and the noise ratio is divided by the noise
threshold of 0.3.

\subsection{Labeling Clusters}\label{subsec:identifying-contexts}
So far we have described how we generate context vectors and how we cluster them.
Each cluster represents one meaning of the target word.
However, for the intended applications, these clusters have to be presented to the users.
Therefore, we need topic words to label and describe them.
We use two methods to generate cluster labels, and both are based on using the average embedding vector,
i.e.\ context vector, which represent the cluster and thus the context.

In our first approach we search for words with the highest cosine-similarity in the IN-embedding space.
These words should represent typical context words for the cluster meaning.
The second approach focuses on the OUT-embedding space, delivering words the network predicts with high
probability as center word for the cluster representative.

    \section{Results and Discussion}\label{sec:results and Discussion}
In this section we qualitatively analyse the clustering results as well as the clustering score.
We show a small excerpt of the clustering results for the selected words.

\subsection{Parameter Score Results}\label{subsec:wikipedia-ambiguity-page-as-reference}
Table~\ref{tab:param-score-results} summarizes the metrics for the best parameter scores for every selected word.
The number of clusters consistently exceed the manually counted meanings in wikipedia pages with homonymous contents,
for every word.

\begin{table}[h]
    \centering
    \caption{The metrics for the top parameter score result for every selected word.}
    \label{tab:param-score-results}
    \begin{tabular}{l|r|r|l|l|r|r|l|l}
        \toprule
        \thead[c]{Word}  & Domains & Meanings & $ \thead[c]{p(c)}$ & $\thead[c]{\epsilon}$ & \thead[c]{$m$} & \thead[c]{$n(c)$} & \thead[c]{$\overline{s}(c)$} & \thead[c]{$noise(c)$} \\
        \midrule
        Mars             & 9       & 86       & 0.165              & 0.21                  & 5              & 104               & -0.40                        & 0.211                 \\
        Maus (en. mouse) & 14      & 51       & 0.209              & 0.2                   & 5              & 78                & -0.35                        & 0.233                 \\
        \makecell[lt]{Kleeblatt \\(en. clover leaf)}            & 6       & 14        & 0.271              & 0.23                  & 5              & 19                & -0.13                        & 0.269                 \\
        Pepsi & 2 & 4 & 0.218 & 0.32 & 5 & 11 & -0.11 & 0.078 \\
        \makecell[lt]{Gabelbein \\(en. wishbone)} & 2 & 2 & 0.596 & 0.2 & 5 & 3 & 0.212 & 0.196 \\
        \makecell[lt]{Datenstruktur \\(en. data structure)} & 1 & 1 & 0.544 & 0.25 & 5 & 8 & 0.302 & 0.084 \\
        \bottomrule
    \end{tabular}
\end{table}

\begin{figure*}[htbp]
    \centering
    \includegraphics[width=\textwidth]{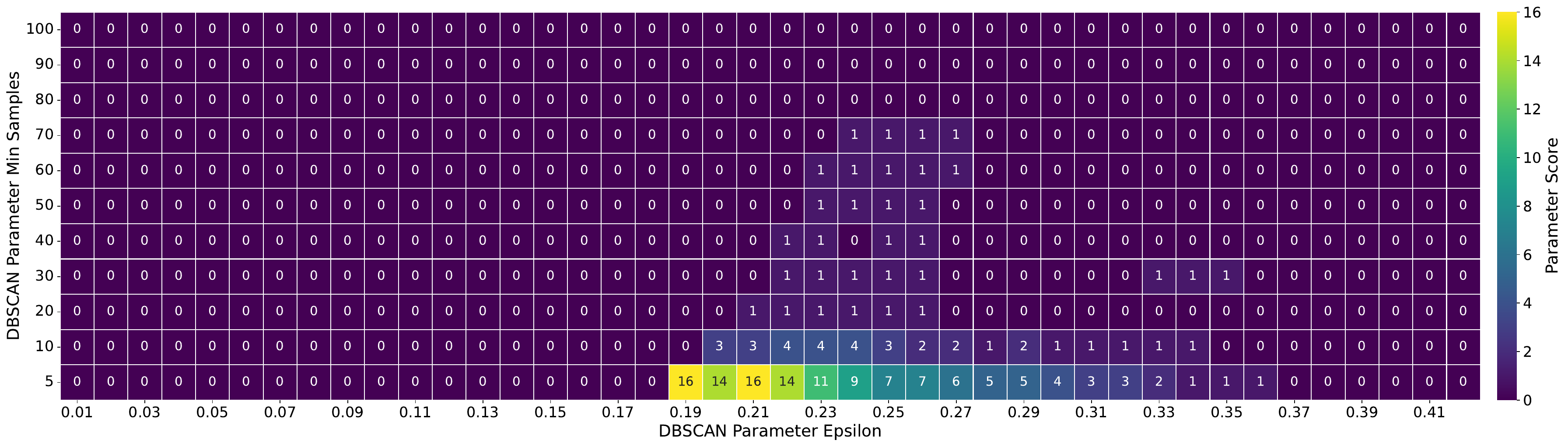}
    \caption{The normalized parameter scores for the clusterings of the word `mars',
        multiplied by 100 for better readability.}
    \label{fig:parameter-score}
\end{figure*}

The parameter score results visible in figure~\ref{fig:parameter-score} clearly propose the parameters $m = 5$ with
$\epsilon = {0.19, 0.21}$ for the word `mars', corresponding to a number of clusters of 110 and 104, respectively.
Clusterings for the word `mars' with $0.43 \geq \epsilon \geq 0.99$ all hold a parameter score of 0,
as the number of clusters for all clusterings equal one, resulting in the default silhouette score of -1,
thus in a parameter score of 0.

\citeauthor{schubertDBSCANRevisitedRevisited2017} (\cite{schubertDBSCANRevisitedRevisited2017}) note that a sudden
increase or decrease in the number of clusters may indicate to have reached the resolution limits of the dataset.
We noticed that our parameter score is highest for the clustering with the last, smallest epsilon value right before
a sudden decrease in the number of clusters.
We interpret this sudden drop, that is also visible in the exemplary heatmap~\ref{fig:cluster_count},
as the resolution limit of the context vectors, and conclude that the proposed parameter score complies to
this heuristic.

\subsection{Context Evaluation}\label{subsec:context-evaluation}
Table~\ref{tab:words} presents the unmodified In2In and In2Out context tokens for the first four clusters of all
selected words by example.
We added the presumed meaning to every cluster.
When looking at the 10 words closest to the mean context vector, it is quite simple to manually assign a single topic
word that describes the words within a single context.
Even if not, we can easily find the meaning of a cluster with a single query in a search engine, simply combining the
target word and a context word.
For example, cluster '2' from the In2In similar words contain the words 'amiga', 'lp', and 'langspielplatte', with 7
dates, ranging from 1981 to 1990, in between.
The single query 'kleeblatt amiga' showed that amiga is a label that had active releases in the given years.

\begin{table}
    \centering
    \caption{The top 3 words from the the first four clusters of the selected words.}
    \label{tab:words}
    \scriptsize
    \begin{tabular}{|c|ll|ll|ll|}
        \toprule
        \emph{Mars} & \multicolumn{2}{c|}{Cluster 1} & \multicolumn{2}{|c|}{Cluster 2} & \multicolumn{2}{|c|}{Cluster 3} \\ \cmidrule{1-7}
        Word & In2In       & In2Out        & In2In        & In2Out      & In2In      & In2Out      \\
        \midrule
        1 & mars & apollo & ausrüsten & \makecell[lt]{maschinen- \\ gewehre}    & commune      & commune         \\
        2 & planet & venus & feldhaubitz & batterie & anctoville & déléguée \\
        3 & jupiter & planet & haubitzen & waffensystem & \makecell[lt]{saint-caprais- \\ de-blaye} & \makecell[lt]{gemeinde- \\ verband}    \\
        \midrule
        Meaning & \multicolumn{2}{c|}{Planets} & \multicolumn{2}{|c|}{Tactical System} & \multicolumn{2}{|c|}{\makecell[ct]{Admin. units \\ `mars' in france}} \\
        \bottomrule

        \toprule
        \emph{Maus} & \multicolumn{2}{c|}{Cluster 1} & \multicolumn{2}{|c|}{Cluster 2} & \multicolumn{2}{|c|}{Cluster 3} \\
        \midrule
        1 & burg & burg & maus       & maus & orthologe & uniprot \\
        2    & \makecell[lt]{burg\_      \\neuhaus} & burghügel & normal & hund & \makecell[lt]{eutel- \\ eostomi}    & orthologe    \\
        3 & \makecell[lt]{burg- \\ anlage} & deidesheim & mauspad & \makecell[lt]{beispiels- \\ weise}    & entrez       & lebewesen         \\
        \midrule
        Meaning & \multicolumn{2}{c|}{Castles} & \multicolumn{2}{|c|}{\makecell[ct]{Computer Periphery \\ and Animals}} & \multicolumn{2}{|c|}{Genetics} \\
        \bottomrule

        \toprule
        \emph{Kleeblatt} & \multicolumn{2}{c|}{Cluster 1} & \multicolumn{2}{|c|}{Cluster 2} & \multicolumn{2}{|c|}{Cluster 3} \\
        \midrule
        1 & balken & kreuz & fcn.de    & kik & amiga & cd \\
        2 & schild & rot & \makecell[lt]{fc\_st. \\ \_pauli}    & pizarro & lp    & box    \\
        3 & hufeisen & wappe & bvb & 2008 & 1982 & amiga \\
        \midrule
        Meaning & \multicolumn{2}{c|}{Heraldry} & \multicolumn{2}{|c|}{Soccer} & \multicolumn{2}{|c|}{\makecell[ct]{`Kleeblatt' LP \\ Amiga record label}} \\
        \bottomrule

        \toprule
        \emph{Pepsi} & \multicolumn{2}{c|}{Cluster 1} & \multicolumn{2}{|c|}{Cluster 2} & \multicolumn{2}{|c|}{Cluster 3} \\
        \midrule
        1    & pepsi    & pepsi & blockflötist      & komponist    & 2001–2003      & 2001–2002      \\
        2 & show & jell-o & \makecell[lt]{volksmusik- \\ forscher} & 1959 & 1998–2001 & 1997–2001 \\
        3 & dallas & live & \makecell[lt]{kultur- \\ manager} & \makecell[lt]{kultur- \\manager} & 2000–2003 & 2002–2005 \\
        \midrule
        Meaning & \multicolumn{2}{c|}{\makecell[ct]{Superbowl \\ Commercial}} & \multicolumn{2}{|c|}{\makecell[ct]{Composer \\ David Lucas}} & \multicolumn{2}{|c|}{\makecell[ct]{Dates of \\ Cola-Wars}} \\
        \bottomrule

        \toprule
        \emph{Gabelbein} & \multicolumn{2}{c|}{Cluster 1} & \multicolumn{2}{|c|}{Cluster 2} & \multicolumn{2}{|c|}{Cluster 3} \\
        \midrule
        1 & \makecell[lt]{becken- \\ knoche} & unterkiefer & radnabe & hydraulisch & auslegen & \makecell[lt]{schraub- \\verbindung} \\
        2 & gelenkfläch & oberkiefer & zahnkranz & verstellbar & reduzieren & steifigkeit \\
        3 & osteoderme & schädel & tauchrohr & antriebswelle & \makecell[lt]{konstruktions- \\ bedingen} & hydraulisch \\
        \midrule
        Meaning & \multicolumn{2}{c|}{Bones} & \multicolumn{2}{|c|}{\makecell[ct]{Telescopic \\ Fork}} & \multicolumn{2}{|c|}{\makecell[ct]{Pros and Cons of \\ Telescopic Forks}} \\
        \bottomrule

        \toprule
        \makecell[lt]{\emph{Daten-} \\ \emph{struktur}} & \multicolumn{2}{c|}{Cluster 1} & \multicolumn{2}{|c|}{Cluster 2} & \multicolumn{2}{|c|}{Cluster 3} \\
        \midrule
        1 & \makecell[lt]{sinnvoller- \\ weise} & \makecell[lt]{implemen- \\ tierung} & verlag & vieweg & iabot & uni-kassel.de \\
        2 & sinnvoll & datenstruktur & lineare & lehrbuch & toter & online \\
        3 & definiert & prozessor & teubner & verlag & \makecell[lt]{internet\_ \\ archive)} & 2008 \\
        \midrule
        Meaning & \multicolumn{2}{c|}{\makecell[ct]{Technical \\ terms}} & \multicolumn{2}{|c|}{Literature} & \multicolumn{2}{|c|}{\makecell[ct]{Outdated \\ References}} \\
        \bottomrule
    \end{tabular}
\end{table}

The first three words of all in2in and in2out clusters can not be clearly divided into
similar (In2IN) and thematic (In2Out) words.
Yet, the meaning, i.e.\ context that every clusters represents are simple to derive.
Only few examples are difficult to decipher, like the second cluster of `pepsi',
where the In2In words would suggest the broader topic `musician', while
the In2Out words clearly suggest a composer, which was simple to find in the 2-word query `komponist pepsi'.
Some tokens are incomplete, such as `Burg Maus' (en.\ castle `Maus') or `beckenknoche' (ref.\ to ger.\ `Beckenknochen'),
while others are special characters, as the pipe `|' in the fourth cluster of `mars'.

    \section{Conclusion}\label{sec:conclusion}
We have shown that different meanings of words can be identified and resolved by clustering context vectors of these words.
Each resulting cluster represents a specific meaning of the word based on specific occurrences in the text corpus.
The main contribution of this paper is an automated approach for DBSCAN parameter selection, based on established,
reviewed, and proven heuristics.
We still see some potential to improve the clustering.
E.g.\ we would like to use an IDF-weighting (inverse document frequency) when computing custer vectors to decrease
the impact of unspecific words.
Furthermore, the work of \citeauthor{huangImprovingWordRepresentations2012}~\cite{huangImprovingWordRepresentations2012}
indicates, that a bigger window size could produce clusters with domain-level, and thus less specific meanings.
However, practical applications require plausible labels for the clusters that can be presented to users.
We think that besides statistical criteria, such as similarity in embedding spaces,
more general linguistic criteria are worth investigating.
Nouns or noun phrases probably make the good labels,
especially noun phrases that contain the original target word seem reasonable choices.
For German, compound words are excellent choices, especially if they contain the original target word,
e.g. \textit{Computermaus} for the target word \textit{Maus}.
The first cluster for the target word \textit{Maus} contains \textit{Burg} as cluster label.
It should be \textit{Burg Maus}, but unfortunately, this entity was not recognized,
as spaCy did not generate a separate word embedding.
We would like to switch to a more elaborate tokenizer, that better identifies named entities and noun phrases, such
as \textit{artificial intelligence}.
If these embeddings also exist for individual tokens and concepts, usually represented as noun phrases,
they could be chosen as cluster labels.



%
    \printbibliography
\end{document}